\documentclass[letterpaper]{article} 
\usepackage{aaai25}  
\usepackage{times}  
\usepackage{helvet}  
\usepackage{courier}  
\usepackage[hyphens]{url}  
\usepackage{graphicx} 
\usepackage{natbib}  
\usepackage{caption} 
\usepackage{algorithm}
\usepackage{algorithmic}

\usepackage{multirow}
\usepackage{adjustbox}
\usepackage{algorithm}
\usepackage{booktabs}
\usepackage{amsmath}
\usepackage{amssymb}
\usepackage{color}

\usepackage{newfloat}
\usepackage{listings}
\DeclareCaptionStyle{ruled}{labelfont=normalfont,labelsep=colon,strut=off} 
\floatstyle{ruled}
\newfloat{listing}{tb}{lst}{}
\floatname{listing}{Listing}
\title{Exploring Unbiased Deepfake Detection via Token-Level Shuffling and Mixing}
\author{
    Xinghe Fu\textsuperscript{\rm 1}\equalcontrib,
    Zhiyuan Yan\textsuperscript{\rm 2}\equalcontrib,
    Taiping Yao\textsuperscript{\rm 2}\thanks{These are corresponding authors.},
    Shen Chen\textsuperscript{\rm 2},
    Xi Li\textsuperscript{\rm 1}\footnotemark[2]
}
\affiliations{
    \textsuperscript{\rm 1}College of Computer Science and Technology, Zhejiang University\\
    \textsuperscript{\rm 2}Youtu Lab, Tencent\\
    xinghefu@zju.edu.cn, \{zhiyuanyan, taipingyao, kobeschen\}@tencent.com, xilizju@zju.edu.cn
}

\begin{document}

\maketitle

\begin{abstract}
The generalization problem is broadly recognized as a critical challenge in detecting deepfakes. 
Most previous work believes that the generalization gap is caused by the differences among various forgery methods. 
However, our investigation reveals that the generalization issue can still occur when forgery-irrelevant factors shift. 
In this work, we identify two biases that detectors may also be prone to overfitting: position bias and content bias, as depicted in Fig.~\ref{fig:bias_vis}.
For the position bias, we observe that detectors are prone to ``lazily" depending on the specific positions within an image (\textit{e.g.}, central regions even no forgery).
As for content bias, we argue that detectors may potentially and mistakenly utilize forgery-unrelated information for detection (\textit{e.g.}, background, and hair). 
To intervene on these biases, we propose two branches for shuffling and mixing with tokens in the latent space of transformers. 
For the shuffling branch, we rearrange the tokens and corresponding position embedding for each image while maintaining the local correlation. 
For the mixing branch, we randomly select and mix the tokens in the latent space between two images with the same label within the mini-batch to recombine the content information.
During the learning process, we align the outputs of detectors from different branches in both feature space and logit space.
Contrastive losses for features and divergence losses for logits are applied to obtain unbiased feature representation and classifiers. 
We demonstrate and verify the effectiveness of our method through extensive experiments on widely used evaluation datasets.
\end{abstract}

\begin{figure}[!t] 
\centering 
\includegraphics[width=\linewidth]{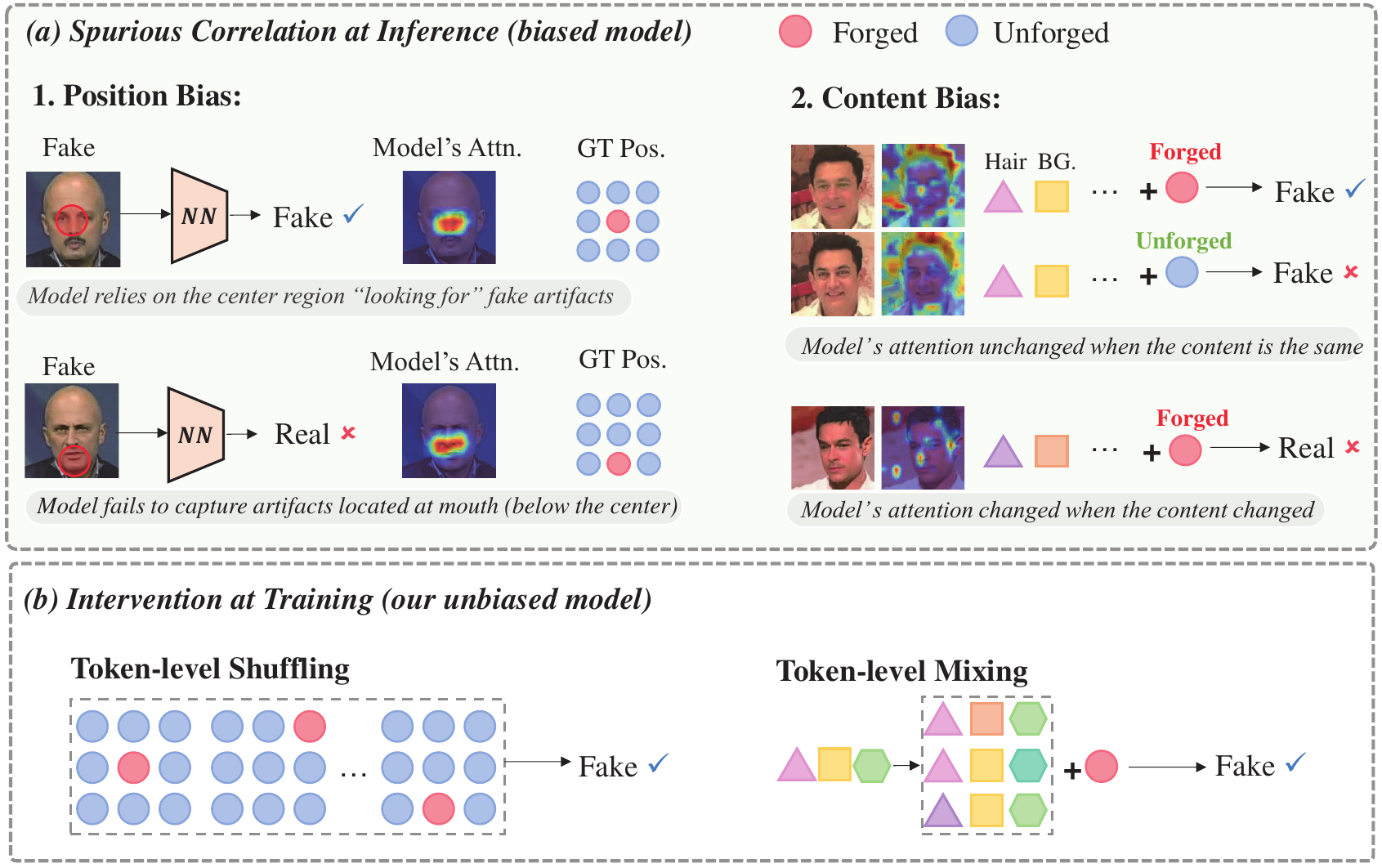} 
\caption{
We present two identified biases in deepfake detection: (1) position bias and (2) content bias. (a) For position bias, we discover that the detector may focus more on the image center, regardless of whether the forgery is present. For content bias, we observe that detectors mistakenly concentrate on the forgery-irrelevant features (\textit{e.g.,} background and hair).
These biases can cause spurious correlations and lead to a biased detector; (b) Our method intervenes on the biases and helps establish a more robust model.
} 
\label{fig:bias_vis} 
\end{figure}

\section{Introduction}
Deepfake technology has become prominent due to its ability to create impressively realistic visual content. However, this technology can also be exploited for harmful purposes, such as violating personal privacy, disseminating false information, and undermining trust in digital media. Considering these potential threats, there is an urgent need to develop a reliable deepfake detection system.

Most earlier deepfake detectors~\cite{li2018exposing,yang219exposing,qian2020thinking, gu2021spatiotemporal} demonstrate effectiveness in the scenario of within-dataset but often falter in cross-dataset scenarios where there is a distribution gap between the training and testing data.  
The prevalent explanation for this generalization problem, as suggested in previous works~\cite{luo2021generalizing,yan2023ucf}, is models' overfitting to specific forgery. These works argue that generalization failure primarily occurs because the forgery methods applied in training and testing data are not identical, leading many subsequent studies~\cite{yan2023transcending, chen2022self, yan2024df40} to address this issue from various methodological perspectives.

In this paper, we discover that the generalization problem persists even when identical forgery techniques are applied. Through our initial investigations, we identify two biases in deepfake detection: position bias and content bias. As illustrated in Fig.~\ref{fig:bias_vis}, we observe that, for position bias, detectors tend to ``lazily" rely on the central region of the image for detection. Concerning content bias, we find that the detector might mistakenly and inadvertently use specific content combinations for detection, \textit{e.g.}, background, hair, or clothes. 
The two key observations inspire us to develop an unbiased detector capable of relying \textit{less on biased information}, thereby creating more robust deepfake detectors.

To address these two biases, we introduce two plug-and-play strategies, namely the \textit{shuffling branch} and \textit{mixing branch}, which simply operate at the token level within ViTs. 
\textbf{Firstly}, in the shuffling branch, we implement a shuffling operation to rearrange the latent token order of a given image and shuffle its corresponding position embedding, to obtain the final shuffled representation. This method aims to disrupt the biased information (\textit{e.g.,} ID~\cite{dong2023implicit}) by reorganizing the spatial position relationship. 
\textbf{Secondly}, in the mixing branch, we exchange a portion of the latent tokens between two images that have the same label within the same mini-batch. The rationale is that swapping certain tokens between two such images (with the same label) should not alter the original decision. 
Specifically, when only a limited number of tokens are interchanged, the remaining tokens should retain the key discriminative features while the content information is recombined.

Our solution presents two potential advantages compared to previous unbiased learning works~\cite{liang2022exploring,yan2023ucf}. 
\textbf{First}, these methods identify the part of the content bias problem and propose a disentanglement framework to overcome this bias through an implicit reconstruction learning process. 
However, they do not address the alleviation of position bias, which may occur even when no face is present in the image. 
In contrast, we propose an explicit unbiased learning method featuring two operations (\textit{i.e.,} shuffling and mixing) designed to disrupt the biased context.
\textbf{Second}, since our proposed methods specifically operate on latent tokens and are lightweight. they can be easily extended to any advanced ViT-based models, including recent state-of-the-art approaches (\textit{e.g.,} CLIP).
On the other hand, previous works~\cite{liang2022exploring,yan2023ucf} are based on a fixed disentanglement framework with extra decoders for reconstruction, which is effective but may not be flexible enough for extension.

Our contributions are summarized as follows:

\begin{itemize}
    \item We identify two critical factors except the forgery specificity that contribute to the generalization problem in deepfake detection: position bias and content bias.

    \item We propose an unbiased deepfake detection approach: UDD, to address position bias and content bias, which involves shuffling and mixing branch and alignment strategies.
    
    \item Extensive experiments show that our framework can outperform the performance of existing state-of-the-art methods in unseen testing datasets, demonstrating its effectiveness in generalization.
\end{itemize}
\section{Related Work}

\subsection{General Deepfake Detection}

The task of deepfake detection presents significant challenges, primarily in capturing the subtle traces of manipulation and enhancing the generalizability of detection models. Prior work in this area has focused on extending the data view, such as analyzing the frequency domain~\cite{qian2020thinking, li2021frequency, li2022wavelet} and leveraging specialized network modules~\cite{zhao2021multi, dang2020detection, song2022face} to capture detailed forgery traces. 

While promising detection performance is achieved, no constraints are presented in these methods that prevent model overfitting and enable the learning of generalized forgery information.
To address the issue of generalization, researchers have employed synthesis and blending techniques in RGB images~\cite{li2020face, chen2022self, shiohara2022detecting, larue2023seeable, li2018exposing}. These methods (\textit{e.g.}, SBIs~\cite{shiohara2022detecting} and SLADD~\cite{chen2022self}) reduce the content disparity between real and fake samples and encourage the model to learn common forgery artifacts for better generalization. 
Some methods~\cite{sun2020aunet, yan2023transcending} also utilize augmentation and synthesis in the latent space to enrich the forgery samples and attain a more robust detection model. Other methods like UCF~\cite{yan2023ucf} and IID~\cite{dong2023implicit} leverage generalized forgery features (like ID difference) in the detection with a specially designed learning framework. 

Unlike previous methods, which focus on general forgery artifacts, we identify the position and content bias for the detection model and directly reduce the bias in the training data.
Specifically, we design the random shuffling and mixing operations and apply them to both real and fake samples to obtain unbiased forgery representation in the proposed learning framework.

\subsection{Unbiased Representation Learning}

The goal of unbiased representation learning methods is to learn feature representations that are invariant to the biases presented in the training data.
Data augmentation plays a crucial role in unbiased representation learning by introducing variability into the training process, which helps to prevent the model from overfitting to spurious correlations in the data. 
Some work~\cite{mitrovic2020representation, ilse2021selecting} theoretically analyzes the effect of data augmentation in learning invariant features from the causal perspective.
Recent advancements in this field have also explored the use of contrastive learning~\cite{williams1992simple, oord2018representation}, where the model is trained to distinguish between similar (positive) and dissimilar (negative) pairs of data points. By doing so, the model learns to focus on the most discriminative features of the data, which are often less susceptible to bias. This approach has been particularly effective in learning representations that are general for various tasks~\cite{rizve2021exploring, ba2024exposing}.

For general deepfake detection, we propose a unified unbiased learning framework to reduce the bias toward spurious correlations like position and content factors.

\section{Method}

\subsection{Overview}

\begin{figure*}[!t] 
\centering 
\includegraphics[width=0.82\textwidth]{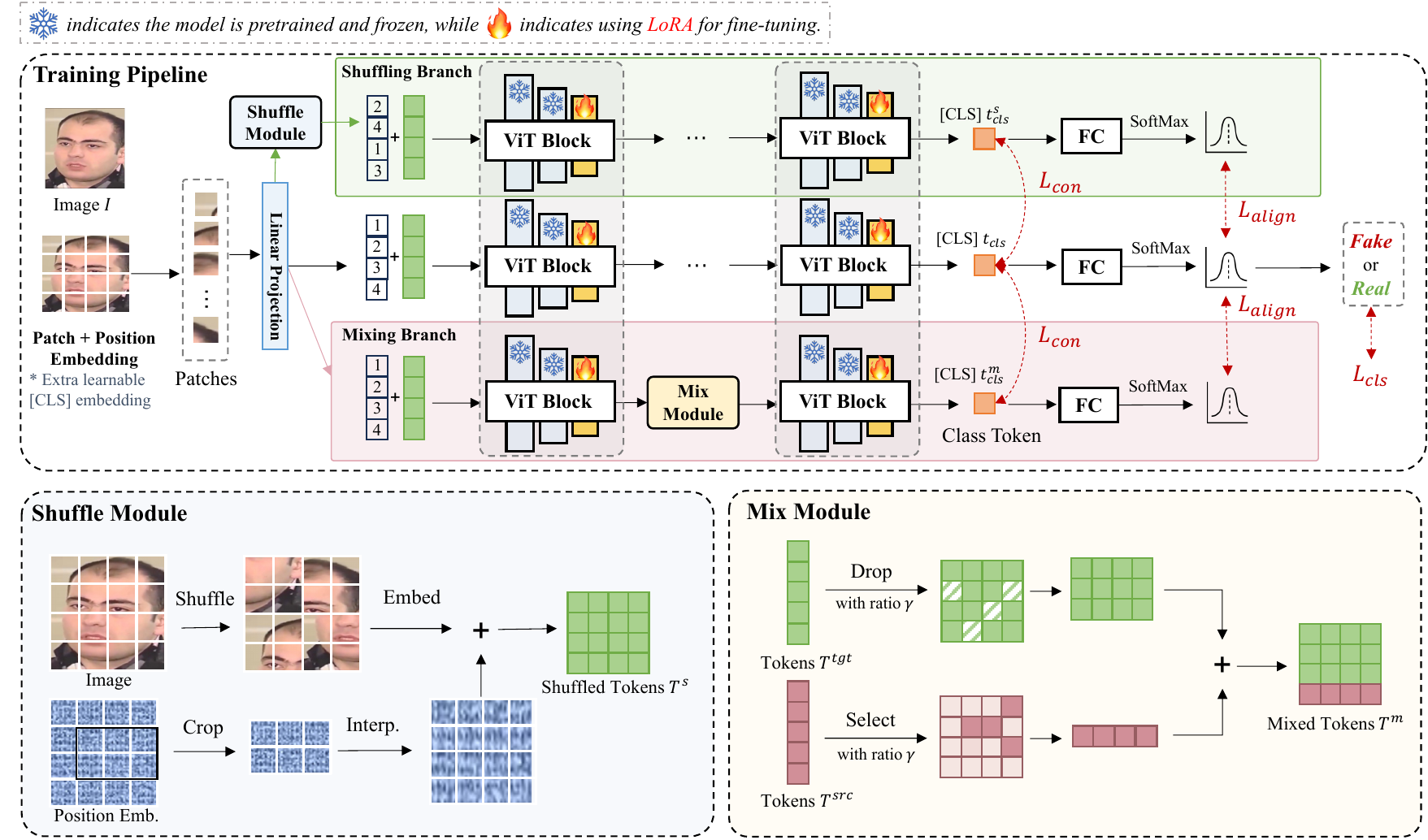} 
\caption{
The overall pipeline of the proposed framework. The input image is sent to the original, shuffling and mixing branch during training. The shuffling branch (S-Branch) introduces the random intervention on position information with the shuffle module at the embedding layer. The mixing branch (M-Branch) introduces the intervention on content information with the mix module between randomly selected blocks. Both operations are applied to token-level representations in the latent space. All branches share the parameters of the network. Contrastive loss $\mathcal{L}_{con}$ and alignment loss $\mathcal{L}_{align}$ are applied over branches to attain unbiased forgery representation and classifier.
} 
\label{fig:pipeline} 
\end{figure*}

The deepfake detection task is commonly modeled as a binary image classification problem. 
The sampling of an image is associated with latent variables of forgery, content, and position, while the sampling of labels is solely related to the forgery latent variable.
Formally, let $X$ denote an image, and $Y$ its corresponding label, where $Y=1$ indicates a forged image and $Y=0$ indicates an authentic one. The sampling of the training set can be formulated as follows:
\begin{equation}
Y\sim p(Y | Z_f), X\sim p(X | Z_f, Z_b), 
\end{equation}
where $Z_f$ represents the forgery variable, and $Z_b = (Z_c, Z_p)$ represents the content and position variables. The label $Y$ is conditioned on the forgery latent variable $Z_f$, while the image $X$ is conditioned on all three latent variables.
This sampling and dataset construction approach may lead to the model capturing spurious correlations (in Fig.~\ref{fig:bias_vis}) between content $Z_c$, position $Z_p$, and labels $Y$, as the distribution of $Z_c$ and $Z_p$ are biased in the training set (\emph{e.g.}, specific forged identity and cropping in the pre-processing). To mitigate the model's bias towards position and content in the learned forgery-related representations, we have designed two branches of augmentation operations based on the token-level latent space of transformer models. Specifically, the operations involve shuffling operations at the embedding layer and mixing operations in the forwarding process. These operations disrupt and recombine the position information and image content, thereby enhancing the randomness in the sampling over these two latent variables and blocking the spurious correlations between them and labels.

The overall framework includes the token-shuffling branch and the token-mixing branch during forwarding, and the feature and logit level alignment loss to attain the unbiased representation and classifier. A causal analysis of the framework is also provided.

\subsection{Token-Shuffling Branch}
\label{s-branch}

To enhance the randomness of the forged position distribution within images and reduce the model's reliance on specific locations, we introduce a shuffling module at the embedding layer of vision transformers (ViTs) to form the token-shuffling branch (in Fig.~\ref{fig:pipeline}). This module comprises two parts: one that performs random rectangular sampling and interpolation on the patch token position encodings, and another that executes blockwise shuffling of the correspondence between patch tokens and their position encodings.

\textbf{Random interpolated position embeddings $pos'_i$.}
To introduce randomness of absolute position and scaling into the position embeddings $pos_i$~\cite{kim2023region, yuan2023densedino}, we reshape the patch position embeddings into a rectangle (\emph{e.g.}, $14\times14$) and perform the following operations. 1) We first sample the aspect ratio $r$, area $S$, and location $(x, y)$ of the rectangle from a uniform distribution. 

The area is kept larger than 30\% of the whole rectangular area of position embeddings. 
2) After sampling, we crop the position embeddings according to the obtained rectangle. 3) The cropped local position embeddings are then interpolated 
and flattened as the random interpolated position embeddings $pos'_i$. 

\textbf{Shuffled patch tokens $e_{\pi (i)}$.}
To enhance the relative position randomness of forgery traces, we introduce random shuffling of patch tokens. Since forgery traces depend on the content of local regions, the shuffling is performed in a blockwise manner and the local correlation within each block is maintained. We reshape the patch tokens into a rectangle and divide it into $s \times s$ blocks. The blocks are then randomly permuted, creating a mapping from the original index $i$ of each patch to a new index $\pi (i)$.

After performing these operations, we add the shuffled patch tokens to the new position embeddings to obtain the token embeddings $t_i \in \mathbb{R}^D$,
\begin{equation}
    t_i = pos'_i + e_{\pi (i)}.
\end{equation}
The shuffling module approximates a $do(\cdot)$-operator on the positional latent variable $Z_p$ and output the token set $T^s=\{t_i\}_{i=1}^{N+1}$ (class token included).
The shuffle module introduces randomness in both relative and absolute position information while preserving the locality of forgery information. This helps the model reduce the bias towards positions.

\subsection{Token-Mixing Branch}
\label{m-branch}

To enhance the randomness of sample content and enable the model to extract unbiased forgery features across different contexts, we designed a token-level content mixing module for the forward process. The mixing module consists of two steps: random dropping target token sets and mixing source and target token sets.

\textbf{Random token dropout.}
For a given set of tokens $T_l$ at layer $l$ during the forward pass, we randomly drop a proportion $\gamma$ of the patch tokens. Let the feature representation of the target image $X$ at layer $l$ be $T_l^{tgt} = \{t_i\}_{i=1}^{N+1}$, with the class token denoted as $t_{cls} = t_{N+1}$. We sample a subset of indices $ID = \{a_i\}$, where $1 \leq a_i \leq N$ and $p(i \in ID) = 1 - \gamma$. The feature representation after token dropout is $T_l^{tgt'} = \{t_i | i \in ID \vee i = N + 1\}$. Assuming that forgery features are local and that tokens have undergone global feature exchange in the latent space, the remaining tokens still represent partial forgery features of the image. 

\textbf{Source \& target mixing.}
For the target features $T_l^{tgt'}$ after dropout, we further select source features $T_l^{src}$ of another sample ($Y^{tgt} = Y^{src}$) from the batch for the mixing operation. As tokens have a global receptive field in the latent space, they can capture global content information. By mixing tokens from different sources in the latent space, we recombine the contextual information of the features, alleviating the risk that forgery features do not rely on some specific facial context. Specifically, we randomly select tokens from the source feature set $T_l^{src}$ to form a subset $T_l^{src'}$, where $|T_l^{src'}| = \gamma N$. We then merge $T_l^{src'}$ and $T_l^{src'}$ to form the mixed token set $T_l^{m}= T_l^{tgt'} \cup T_l^{src'}$
and use this token set for the subsequent forwarding process.

We implement the token-mixing branch by inserting the mixing module at randomly selected layers in the forwarding process. In the token-mixing branch, we achieve a random combination of content contexts, approximating a $do(\cdot)$-operator on the content latent variable $Z_c$, introducing content randomness. 

\subsection{Overall Framework}
\label{framework}

Building upon the two operation branches for position and image content intervention, we have designed an unbiased forgery representation learning framework including the tuning architecture and loss functions. In the framework, vision transformers are trained to capture the causal correlation between input images $X \in \mathbb{R}^{3\times H \times W}$ and forgery labels $Y \in \{0, 1\}$ with less bias towards positions and image content in the three-branch data flows. 

\textbf{Architecture.}
It is important to retain and utilize the knowledge for deepfake detection tasks. Therefore, we freeze the backbone parameters and introduce only a small number of learnable parameters (\emph{e.g.}, LoRA~\cite{hu2021lora}). For forged images, the model is required to focus on the forged regions within the attention mechanism and encode forgery-related information during the forward process. Hence, we incorporate learnable parameters in both the multi-head attention and MLP layers. Taking the multi-head attention in transformers as $MSA(\cdot;\{W_a\})$ and the MLP layer in transformers as $MLP(\cdot;\{W_m\})$, where $W_a$ and $W_m$ represent arbitrary pre-trained linear projection matrix in $MSA$ and $MLP$, the forwarding process of a transformer block with additional learnable parameters is as follows:

\begin{align}
    T_l'&= MSA(T_l;\{W_a + \Delta W_a\}) + T_l,\\
    T_{l+1} &= MLP(T_l';\{W_m + \Delta W_m\}) + T_l',   
\end{align}

where $T_l \in \mathbb{R}^{N\times L \times D}$ denotes the input tensor of the $l$-th transformer block,  $\Delta W_a$ and $\Delta W_m$ denote the low-rank matrix. Suppose that the rank of $\Delta W \in \mathbb{R}^{d_1 \times d_2}$ ($\Delta W_a$ or $\Delta W_m$) is $r$, we decompose $\Delta W$ as $\Delta W = A B^T$ and take low-rank matrices $A \in \mathbb{R}^{d_1 \times r}$ and $B \in \mathbb{R}^{d_2 \times r}$ (as in LoRA) as learning parameters during training. It turns out that the design not only achieves parameter-efficient and stable training but also alleviates the overfitting risk in the detection.

\textbf{Learning Objective.}
For the loss function design, we align the output from the original branch with the token-shuffling and token-mixing branches on two levels (\textbf{feature level} and \textbf{logit level}) to obtain unbiased feature representations and classifiers.
At the feature level, we employ the contrastive loss function (as in SimCLR~\cite{chen2020simple}), aligning the class token $t_{cls}$ with $t_{cls}^s$ and $t_{cls}^m$ as positive sample pairs after mapping through a three-layer MLP projector $g(\cdot)$. The feature-level contrastive loss is as follows:
\begin{eqnarray}
\mathcal{L}_{c}(t', t'^+) = -\log \frac{e^{\text{sim}(t', t'^+) / \tau}}{e^{\text{sim}(t', t'^+)/ \tau}+\sum_{t'^-} e^{\text{sim}(t', t'^-) / \tau}}, \\
\mathcal{L}_{con} = \mathcal{L}_{c}(g(t_{cls}), g(t_{cls}^s)) + \mathcal{L}_{c}(g(t_{cls}), g(t_{cls}^m)),
\end{eqnarray}
where $t'=g(t_{cls})$ and $\text{sim}(x, y) = x^Ty/||x||_2||y||_2$. We take other samples within a batch after $g(\cdot)$ as negative samples in the contrastive loss.
Since $t_{cls}$, $t_{cls}^s$, and $t_{cls}^m$ contain similar forgery features, this loss enables aligned feature representations. At the logit level, we impose constraints on the prediction probabilities using Jensen–Shannon divergence here and align the predicted logits after the classifier:
\begin{equation}
    \mathcal{L}_{align} = D_{\text{JS}}(P_{Y|t_{cls}}||P_{Y|t_{cls}^s}) + D_{\text{JS}}(P_{Y|t_{cls}}||P_{Y|t_{cls}^m})
\end{equation}
The overall loss function is as follows (given that $\mathcal{L}_{ce}$ is the cross-entropy loss for classification):

\begin{equation}
\mathcal{L}_{total} = \mathcal{L}_{ce} + \lambda_{1}\mathcal{L}_{con} + \lambda_{2}\mathcal{L}_{align}
\end{equation}

\begin{figure}[t]
\centering 
\includegraphics[width=0.88\linewidth]{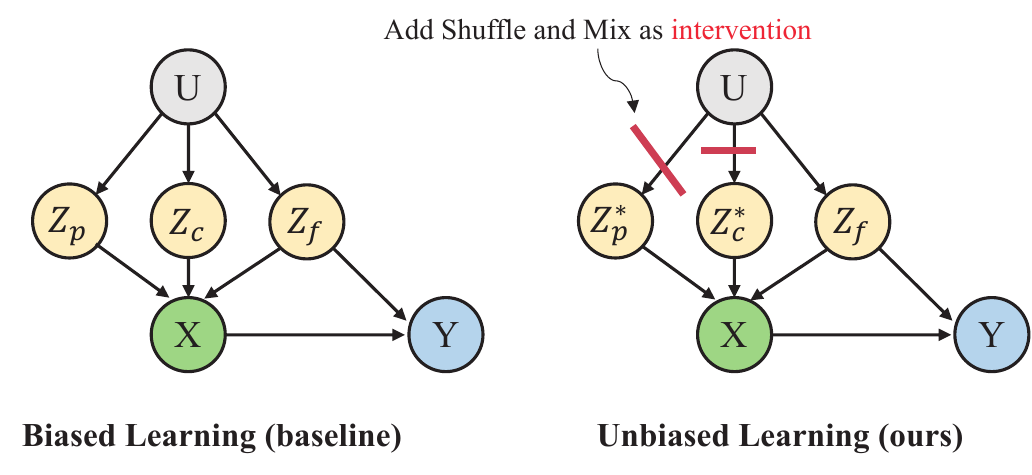} 
\caption{
Causal graph for illustrating the proposed framework (the right) versus the baseline (the left). 
Within the graph, $X$ and $Y$ are observed. 
The unobserved confounder $U$ causes a backdoor path from the position ($Z_p$) and content ($Z_c$) variables to the label $Y$.
Differing from the baseline, our proposed unbiased learning method performs an intervention that blocks the backdoor paths for training an unbiased detector.
} 
\label{fig:causal_graph} 
\end{figure}

\subsection{Causal Analysis} 
\label{causal}

In this subsection, we aim to perform an analysis from the view of causal learning to illustrate the effectiveness of the proposed strategies.
In the previous sections, we show the existence of \textit{spurious correlations} 
(see Fig.~\ref{fig:bias_vis}) when naively training a model to detect deepfakes.
From the view of causal learning, our goal is to design an intervention to mitigate such spurious correlations in the original data distribution. 
In this work, we observe two main factors that contribute to this spurious correlation (\textit{i.e.,} position bias and content bias).
Thus, it is important to disentangle these forgery-unrelated biases (confounders) from the forgery-related causal features. 

The \textit{causal graph} is shown in Fig.~\ref{fig:causal_graph} for unbiased deepfake detection.
We denote the unobserved \textit{confounder} as $U$, which produces forgery-irrelevant nuisance factors (\textit{i.e.,} $Z_p$ and $Z_c$) and forgery-related (causal) feature (\textit{i.e.,} $Z_f$).
For convenience, we use $Z_b$ to represent the forgery-irrelevant bias (including $Z_p$ and $Z_c$ in our case).
The correlation between $Z_b$ and $Y$ is mediated through $U$, forming a backdoor\footnote{With a backdoor, classifiers trained directly on $X$ and $Y$ will not be causal with $Z_f$.} path $X \leftarrow Z_b \leftarrow U \rightarrow Z_f \rightarrow Y$. The existence of this backdoor path leads to a possible bias towards $Z_b$ during training, which hinders the learning of correct causal path $X \leftarrow Z_f \rightarrow Y$.
To achieve unbiased deepfake detection, we need to intervene on $Z_b$ and break the backdoor path.

Here, we use the \textit{``do"-operator} \footnote{The ``do"-operator formalizes the process of intervening in a system. In contrast to conventional statistical methods that demonstrate correlation, the ``do"-operator enables us to model the consequences of actively manipulating a variable.} 
(formally expressed as $do(X=x)$) 
to denote performing a treatment $z$ on the variable $Z_b$ which makes $Z_b$ independent from the causal variable $Z_f$ ($Z_b \! \perp \!\!\! \perp Z_f$) \cite{pearl}. 
We denote the causality from $X$ to $Y$ predicted by networks to be $P_{\theta}(Y|X)$, which is the treatment effect of an input image $X$ on label $Y$.

Removing the backdoor, the correlation learned in the framework is now equal to the causality, \textit{i.e.,} $P_{\theta}(Y|X) = P_{\theta}(Y|Z_f, Z_b) = P_{\theta}(Y|Z_f, do(Z_b))$. 
The proposed shuffling and mixing operations can be regarded as two different interventions added at the token level to the image. This enables the model to remove the spurious correlation (position bias and content bias) and learn more general forgery features for deepfake detection.

\begin{table*}[tbp]
    \centering
    \caption{Comparison with previous methods. We report both frame-level and video-level AUC (\%) of our models trained on FF++ (c23) and compare the results with previous SOTA methods. 
    Methods with * are our reproduction results using the released models. \textbf{Bold} and \underline{underline} indicate the best and the second-best results.
    We report and cite the results of other methods from their original papers or DeepfakeBench~\cite{deepfakebench}.
    }
    \label{tab:cross-eval}
    \begin{tabular}{c|c|c c c c| c |c| c c c c }
    \toprule
       Type  & Method & CDF & DFDCP & DFDC & DFD & Type  & Method & CDF & DFDCP & DFDC & DFD \\
       \midrule
       \multirow{14}{*}{Frame-Level}
         & Xception & 73.7 & 73.7 & 70.8 & 81.6
         & \multirow{14}{*}{Video-Level} & Xception & 81.6 & 74.2 & 73.2 & 89.6\\
         & Efficient-b4 & 74.9 & 72.8 & 69.6 & 81.5 
         & & Efficient-b4 & 80.8 & 68.0 & 72.4 & 86.2\\
         & FWA & 66.8 & 63.7 & 61.3 & 74.0
         & & LipForensics & 82.4 & - & 73.5 & -\\
         & Face X-ray & 67.9 & 69.4 & 63.3 & 76.7 
         & & FTCN & 86.9 & 74.0 & 71.0 & 94.4\\
         & RECCE & 73.2 & 74.2 & 71.3 & 81.2 
         & & RECCE & 82.3 & 73.4 & 69.6 & 89.1\\
         
         & F3-Net & 73.5 & 73.5 & 70.2 & 79.8
         & & F3-Net & 78.9 & 74.9 & 71.8 & 84.4 \\
         & SPSL & 76.5 & 74.1 & 70.4 & 81.2
         & & PCL+I2G & 90.0 & 74.4 &  67.5 & -\\
         
         & SRM & 75.5 & 74.1 & 70.0 & 81.2 
         & & SBIs* & 90.6 & \underline{87.7} & 75.2 & 88.2\\
         & UCF & 73.5 & 73.5 & 70.2 & 79.8
         & & UCF & 83.7 & 74.2 & 77.0 & 86.7 \\
         & IID & 83.8 & \underline{81.2} & - & - 
         & & SeeABLE & 87.3 & 86.3 & 75.9 & -\\
         & ICT & \underline{85.7} & - & - & 84.1
         & & UIA-ViT & 82.4 & 75.8 & - & \underline{94.7} \\
         & LSDA & 83.0 & 81.5 & \underline{73.6} & \underline{88.0}
         & & TALL++ & \underline{92.0} & - & \underline{78.5} & - \\
         & ViT-B (IN21k) & 75.0 & 75.6 & 73.4 & 86.4 
         & & ViT-B (IN21k) & 81.7 & 77.7 & 76.3 & 89.5\\
         & ViT-B (CLIP) & 81.7 & 80.2 & 73.5 & 86.6
         & & ViT-B (CLIP) & 88.4 & 82.5 & 76.1 & 90.0 \\
         \cmidrule{2-6}
         \cmidrule{8-12}
         & UDD (Ours) & \textbf{86.9} & \textbf{85.6} & \textbf{75.8} & \textbf{91.0}
         & & UDD (Ours) & \textbf{93.1} & \textbf{88.1} & \textbf{81.2} & \textbf{95.5}\\
    \bottomrule
    \end{tabular}

\end{table*}

\section{Experiments}

\subsection{Setup}

\textbf{Datasets.} For comprehensively assess the proposed method, we utilize five widely-used public datasets FaceForensics++ (FF++)~\cite{rossler2019faceforensics++}, Celeb-DF (CDF)~\cite{li2019celeb}, DFDC~\cite{dfdc}, DFDC-Preview (DFDCP)~\cite{dfdcp}, and DFD~\cite{dfd} in our experiments, following previous works~\cite{mohseni2020self, yan2023deepfakebench}. FF++ dataset is the most widely used dataset in deepfake detection tasks. We use the training split of FF++~\cite{rossler2019faceforensics++} as the training set. For the cross-dataset evaluation, we test our model on datasets other than FF++. For the robustness evaluation, we test our model on the test split of FF++.

\textbf{Implementation details.}
We utilize the ViT-B model as the backbone network of detectors. The backbone is initialized with the pre-trained weights from the vision encoder of CLIP~\cite{radford2021learning} by default. We evenly sample 8 frames~\cite{mohseni2020self} from the training videos of FF++ (c23) to form the training set.
For S-Branch, we divide the token map into $2\times2$ blocks in shuffling. For M-Branch, the mixing ratio $r$ is set to 0.3. The rank for $\Delta W$ is set to 4. The hyperparameters $\tau$, $\lambda_1$, and $\lambda_2$ in loss functions are set to 0.1, 0.1, and 0.1 in the training. S-Branch and M-Branch are not applied at inference. 

\textbf{Metrics.} We utilize area under receiver operating characteristic curve (AUC) scores for empirical evaluation in experiments. Frame-level AUC is calculated based on the predicted scores of frame inputs. We calculate the video-level AUC with averaged predicted scores of 32 frames sampled from a video. Video-level AUC scores are reported in experiments unless otherwise specified.

\begin{figure*}[t!]
    \centering
    \includegraphics[width=\textwidth]{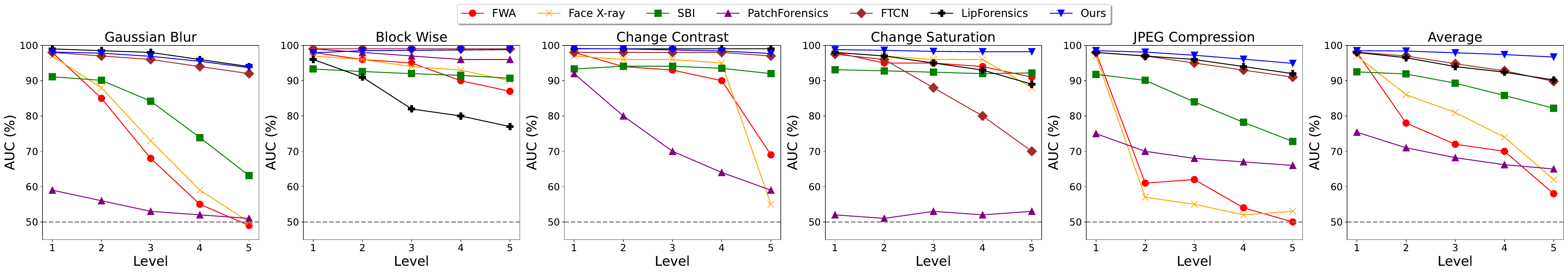}
    \caption{The results of robustness evaluation on the test set of FF++ (c23). Video-level AUC (\%) is reported under five different types of perturbations following~\cite{jiang2020deeperforensics}. Our method is more robust than previous methods across corruptions.}
    \label{fig:robust}
\end{figure*}

\subsection{Comparison with Existing Methods}

\textbf{Generalization evaluation.}
In the cross-dataset evaluation part, models are trained on the FF++ (c23) dataset and evaluated on CDF, DFDCP, DFDC, and DFD datasets. We report both frame-level and video-level AUC scores in Table~\ref{tab:cross-eval} and compare our methods with previous state-of-the-art methods, \textit{e.g.}, LSDA~\cite{yan2023transcending} and TALL++~\cite{xu2024learning}. 
To achieve great improvement in generalization, we focus on reducing the overfitting or bias to forgery-irrelated information like position or content. Our method achieves state-of-the-art results in both frame-level and video-level AUC scores.
For frame-level AUC scores, our method achieves 86.9\% on CDF and outperforms ViT-based state-of-the-arts method ICT~\cite{dong2022protecting} and frequency-based methods~\cite{qian2020thinking, li2021frequency, luo2021generalizing}. Our method also achieves improvement on DFDCP by 4.4\% (from 81.2\% to 85.6\%), DFDC by 2.3\% (from 73.5\% to 75.8\%), and DFD (from 88.0\% to 91.0\%), outperforming RECCE~\cite{cao2022end}, IID~\cite{huang2023implicit} and LSDA~\cite{yan2023transcending}. For video-level AUC scores, our method surpasses previous state-of-the-art methods with image augmentation and synthesis \textit{i.e.}, PCL+I2G~\cite{zhao2021learning} and SeeABLE~\cite{larue2023seeable} on CDF (from 90.0\% to 93.1\%) and DFDCP (from 87.7\% to 88.1\%). On DFDC and DFD datasets, our method outperforms the second-best method 
TALL++(from 78.5\% to 81.2\%) and UIA-ViT~\cite{zhuang2022uia} (from 94.7\% to 95.5\%). The comparison results demonstrate the generalization capability of the proposed method and verify the importance of unbiased forgery representation learning.

\textbf{Robustness evaluation.} To verify the robustness of the proposed method, different types of perturbations from~\cite{jiang2020deeperforensics} are applied to the test set of FF++ (c23). Video-level AUC scores are reported at different perturbation levels (level 5 is the most severe) in Fig.~\ref{fig:robust}. Previous methods~\cite{haliassos2021lips, zheng2021exploring, chai2020makes} achieve great detection performance with full-training. However, these methods may capture abundant low-level forgery clues that are sensitive to perturbations like noise and compression Furthermore, CNN-based methods~\cite{tan2019efficientnet} are obstructed by some specific perturbations like blockwise noise from Fig.~\ref{fig:robust}, while transformer models can easily overlook these noises in the attention mechanism. Our method achieves the highest averaged AUC scores in the robustness evaluation and is less sensitive to severe perturbations of different types than previous state-of-the-art methods.

\begin{figure*}[!tp]
    \centering
    \includegraphics[width=\textwidth]{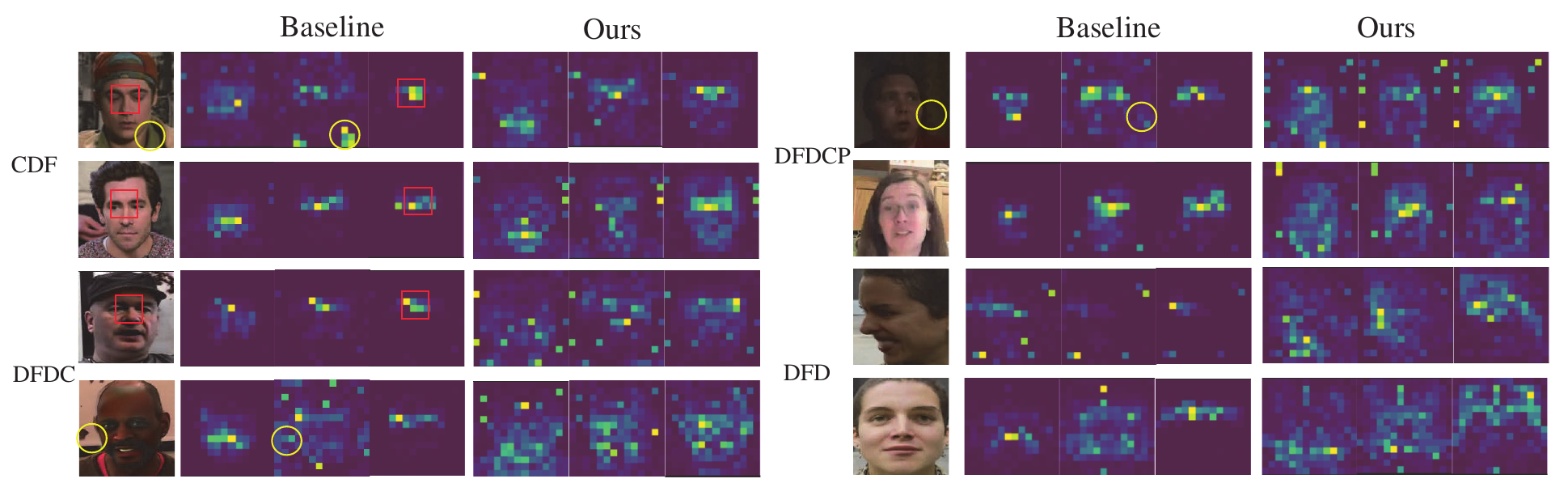}
    \caption{The visualization of multi-head attentions. For visualization, we select attention maps with clear activation in the last layer of ViT models. Squares and circles indicate the position bias and content bias in the baseline model.}
    \label{fig:vis_attn}
\end{figure*}

\subsection{Ablation Study}

\textbf{Ablation of components.} We study the generalization effect of the proposed components with cross-dataset evaluation in Table~\ref{tab:comp}. 
Both the shuffling branch (S-Branch) and the mixing branch (M-Branch) play a key role in achieving state-of-the-art generalization results. Quantitatively, video-level AUC scores improve from 90.70\% (without both) to 93.13\% (with both) on CDF, and from 78.25\% (without both) to 81.21\% (with both) on DFDC. The results indicate that alleviating the position and content bias during training is important to the generalization capability of the detection model.  
Furthermore, The AUC scores of our method drop by over 1\% on CDF, DFDCP, and DFDC when the contrastive loss $\mathcal{L}_{con}$ or the alignment loss $\mathcal{L}_{align}$ is not applied. 
Our method exhibits a significant advantage over the baseline model in the generalization capability when these components are applied (at least 4.72\% improvement).

\begin{table}[htbp]
    \centering
    \caption{Ablation of the proposed components. Video-level AUC (\%) is reported for comparison.}
    \label{tab:comp}

    \begin{tabular}{c|c c c c}
        \toprule
       Model  &  CDF & DFDCP & DFDC & DFD \\
       \midrule
        Baseline & 88.41 & 82.46 & 76.07 & 90.04 \\
        \midrule
        +$\Delta W$ & 90.70 & 85.97 & 78.25 & 93.08\\
        w/o S-Branch & 91.25 & 87.30 & 81.02 & 94.95\\
        w/o M-Branch & 90.51 & 87.68 & 80.33 & 92.25\\
        w/o $\mathcal{L}_{con}$ & 91.20 & 86.35 & 79.74 & 94.03\\
        w/o $\mathcal{L}_{align}$ & 90.52 & 87.01 & 80.18 & 94.62 \\
        \midrule
        Ours & \textbf{93.13} & \textbf{88.11} & \textbf{81.21} & \textbf{95.51}\\
         \bottomrule
         
    \end{tabular}

\end{table}

\begin{table}[tp]
    \centering
    \caption{Ablation of pre-trained weights. Video-level AUC (\%) is reported for comparison. }
    \label{tab:pretrain}
    \begin{tabular}{c|c c c c}
        \toprule
       Model & CDF & DFDCP & DFDC & DFD \\
       \midrule
        ViT-B (IN21k) & 81.66 & 77.72 & 76.30 & 89.82\\
        +Ours & \textbf{88.56} & \textbf{83.84} & \textbf{76.50} & \textbf{92.73} \\
        \midrule
        ViT-B (CLIP) & 88.41 & 82.46 & 76.07 & 90.04\\
        +Ours & \textbf{93.13} & \textbf{88.11} & \textbf{81.21} & \textbf{95.51}\\
        \bottomrule
    \end{tabular}
\end{table}

\begin{table}[t]
    \centering
    \caption{Ablation of mixing stages. Video-level AUC (\%) is reported for comparison. ``Before first, early, mid, and late" indicate different positions where the mix module is inserted in the token-mixing branch. 
    }
    \label{tab:stage}
    \begin{tabular}{c|c c c c}
        \toprule
       Model & CDF & DFDC & DFD & Avg. \\
       \midrule
        Before First & 92.44 & 76.82 & 92.25 & 87.17\\
        Early & \textbf{92.55} & 78.50 & 94.72 & 88.59\\
        Mid & 91.25 & \textbf{81.02} & \textbf{94.95} & \textbf{89.07} \\
        Late & 91.02 & 79.06 & 93.44 & 87.84\\
        \bottomrule
    \end{tabular}
\end{table}

\textbf{Ablation of pre-trained weights.} We study the influence of pre-trained weights for the backbone networks in Table~\ref{tab:pretrain}. The results demonstrate that the selection of the pre-trained model is also significant to the generalization capability in deepfake detection. Pre-trained weights from CLIP show great performance improvements on some datasets (\textit{e.g.}, CDF) compared with weights trained on ImageNet (IN21k). 
However, the improvement brought by our method is insensitive to the selection of backbones from Table~\ref{tab:pretrain}. The AUC scores increase from 81.66\% (w/o ours) to 88.56\% (w/ ours) on CDF and from 77.72\% (w/o ours) to 83.84\% (w/ ours) on DFDCP when using ImageNet pre-trained weights.

\textbf{Ablation of mixing stages.} To identify the appropriate position to insert mixing modules in the M-Branch, we conduct the ablation on the M-Branch and the results are shown in Table~\ref{tab:stage}. We divide the forwarding process (from the 1st to the 2nd-last block) of the ViT-B model into three stages. 
The mixing module is randomly applied after one of the blocks in a specified stage during forwarding. 
Since the content information contained in image patches is limited, mixing before the first block (the first row in Table~\ref{tab:stage}) cannot effectively introduce the content information from other images and achieves sub-optimal generalization performance (the lowest averaged AUC score 87.17\%). In the forwarding process, tokens perceive the information over the whole image with the attention mechanism and capture high-level content semantics in the later blocks. However, mixing in the late stage shows less effect on the final output.
Using blocks in the mid-stage for mixing achieves the best overall performance (from 87.17\% to 89.07\% in the avg. AUC).

\textbf{Visualization of attention maps.} We visualize the attention map for fake samples (class tokens as queries) of the last attention layer in transformers for better viewing of the position and content bias in generalized deepfake detection. To better visually demonstrate, we selected attention maps with clear responses from three heads for display and compared our model with the baseline in Fig.~\ref{fig:vis_attn}. Red and yellow circles highlight the position and content bias in the attention maps of the baseline model. It is observed that the baseline model relies on some specific regions (\textit{e.g.}, upper center) and content information (\textit{e.g.}, clothes or background) for detection. At the same time, our method captures diverse forgery artifacts in the attention maps and relies less on biases. 

\section{Conclusion}

This paper identifies position and content biases that can cause the generalization problem in deepfake detection. 
To mitigate these biases, we introduce UDD, a plug-and-play detection approach that learns unbiased forgery representations from a causal perspective. To achieve the intervention on the position and content, we propose the token-level shuffling and mixing branches for transformers. We also introduce feature-level contrastive loss and logit-level alignment loss to acquire unbiased feature representation and classifiers. Extensive experiments confirm the generalization capability and robustness of the proposed approach. 

\section*{Acknowledgements}

This work is supported in part by National Natural Science Foundation of China under Grant No.62441602, National Science Foundation for Distinguished Young Scholars under Grant 62225605, Zhejiang Provincial Natural Science Foundation of China under Grant LD24F020016, Project 12326608 supported by NSFC, ``Pioneer" and ``Leading Goose" R\&D Program of Zhejiang (No. 2024C01020), and the Fundamental Research Funds for the Central Universities.

\bibliography{aaai25}

\appendix

\section{Training Setup}

\begin{table}[tbp]
    \centering
    \caption{Training setup of our framework used in experiments.}
    \label{tab:setup}
    \begin{tabular}{c|c}
    \toprule
        Terms & Parameters \\
    \midrule
       Backbone  &  ViT-b/16 \\
       Learning rate  & 5e-4\\
       Batch size & 64 \\
       Epochs & 100 \\
       Optimizer & AdamW\\
       $\beta_1$ & 0.9 \\
       $\beta_2$ & 0.999 \\
       $\epsilon$ & 1e-3 \\
       Weight decay & 1e-2\\
       Warmup Scheduler & Linear\\
       Warmup Epochs & 5\\
       Scheduler & CosineAnnealing\\
       Shuffling blocks $s\times s$ & $2\times2$\\
       Aspect ratio $\alpha$ & $(3/4, 4/3)$\\
       Area $S$ & [60, 196]\\
       Left $x$ & $[0, 14)$\\
       Top $y$ & $[0, 14)$\\
       Mixing ratio $\gamma$ & 0.3 \\
       Temperature $\tau$ & 0.1 \\
       Loss weight $\lambda_1$ & 0.1\\
       Loss weight $\lambda_2$ & 0.1\\
    \bottomrule
       
    \end{tabular}
    
\end{table}

We provide the detailed training setup used in the experiments of our method for reproducibility in Table~\ref{tab:setup}. The \textbf{pre-processing} of the datasets follows publicly accessible DeepfakeBench~\cite{yan2023deepfakebench}. The optimizer used in the experiments is AdamW and the corresponding parameters (\textit{i.e.}, $\beta_1$, $\beta_2$, $\epsilon$, and weight decay) are provided in the table. We use a warmup linear scheduler in the first 5 epochs and a cosine annealing scheduler for the rest epochs during training. The parameters of the shuffle module in S-Branch including the number of shuffling blocks $s\times s$ and parameters (\textit{i.e.}, $\alpha$, $S$, $x$ and $y$) for the interpolation of position embeddings as given in Table~\ref{tab:setup}. The mixing ratio $\gamma$ of the mix module in the M-Branch is set to 0.3 by default. The temperature $\tau$ is set to 0.1, and the loss weights $\lambda_1$ and $\lambda_2$ are both set to 0.1.
\section{Ablation}

\subsection{Ablation on Hyperparameter $\lambda_1$ for $\mathcal{L}_{con}$}

We present an ablation study conducted to select the appropriate loss weight $\lambda_1$ for $\mathcal{L}_{con}$ with frame-level AUC scores on CDF (in Table~\ref{tab:abl1}). We experiment with different values of $\lambda_1$, specifically 0, 0.1, 0.5, and 1. Other hyperparameters remain the same as in the setup. The results indicate that the model achieves the best performance when $\lambda_1$ is set to 0.1 (from 85.59\% to 86.74\% compared with $\lambda_1 = 0$). Larger values for $\lambda_1$ encourage the learning of invariant representations for images and have a negative impact on the learning of general forgery features. Therefore, a relatively small value for $\lambda_1$ is enough for feature-level alignment.

\begin{table}[htbp]
    \centering
    \caption{Ablation results on loss weight $\lambda_1$. Frame-level AUC (\%) scores on CDF are reported for comparison.}
    \label{tab:abl1}
    \begin{tabular}{c|c c c c}
        \toprule
       $\lambda_1$  &  0 & 0.1 & 0.5 & 1\\
       \midrule
         Ours & 85.59  &\textbf{86.74} & 85.97 & 83.55\\
         \bottomrule
    \end{tabular}
\end{table}

\subsection{Ablation on Hyperparameter $\lambda_2$ for $\mathcal{L}_{align}$}

We ablate the selection of different values for $\lambda_2$ to determine the appropriate value. As shown in Table~\ref{tab:abl2}, frame-level AUC scores on CDF are reported for comparison (other settings remain the same). It is observed that the model achieves the best performance (86.74\% AUC score) when $\lambda_2$ is set to 0.1. However, the performance starts to degenerate when larger values (like 1.0) are adopted. This may be caused by the incompatibility between the loss weight $\lambda_2$ for JS divergence and the learning rate. Overall, the performance is less sensitive to $\lambda_2$ compared with $\lambda_1$. We select a proper value for $\lambda_2$ to achieve the alignment in the logit space.

\begin{table}[htbp]
    \centering
    \caption{Ablation results on loss weight $\lambda_2$. Frame-level AUC (\%) scores on CDF are reported for comparison.}
    \label{tab:abl2}
    \begin{tabular}{c|c c c c}
        \toprule
       $\lambda_1$  &  0 & 0.1 & 0.5 & 1\\
       \midrule
        Ours & 85.75 & \textbf{86.74} & 86.30 & 85.46\\
         \bottomrule
    \end{tabular}
    
\end{table}

\subsection{Ablation on Temperature $\tau$ in $\mathcal{L}_{con}$}

The temperature $\tau$ has similar effect as $\lambda_1$ in $\mathcal{L}_{con}$. We also ablate the selection of values for $\tau$ in Table.~\ref{tab:abl_tau} and frame-level AUC scores on CDF are reported. Our model achieves optimal generalization performance when $\tau$ is set to 0.1. The performance drop is observed when larger values are adopted since larger values for $\tau$ also negatively influence the learning of general forgery artifacts. We set $\tau$ to 0.1 in other experiments.

\begin{table}[htbp]
    \centering
    \caption{Ablation results on temperature $\tau$. Frame-level AUC (\%) scores on CDF are reported for comparison.}
    \label{tab:abl_tau}
    \begin{tabular}{c|c c c c}
        \toprule
       $\tau$  &  0.05 & 0.1 & 0.15 & 0.2\\
       \midrule
        Ours & 85.87  &\textbf{86.74} & 85.36 & 85.46\\
         \bottomrule
    \end{tabular}
    
\end{table}

\subsection{Ablation on Shuffling Blocks $s \times s$}

\begin{figure}
    \centering
    \includegraphics[width=\linewidth]{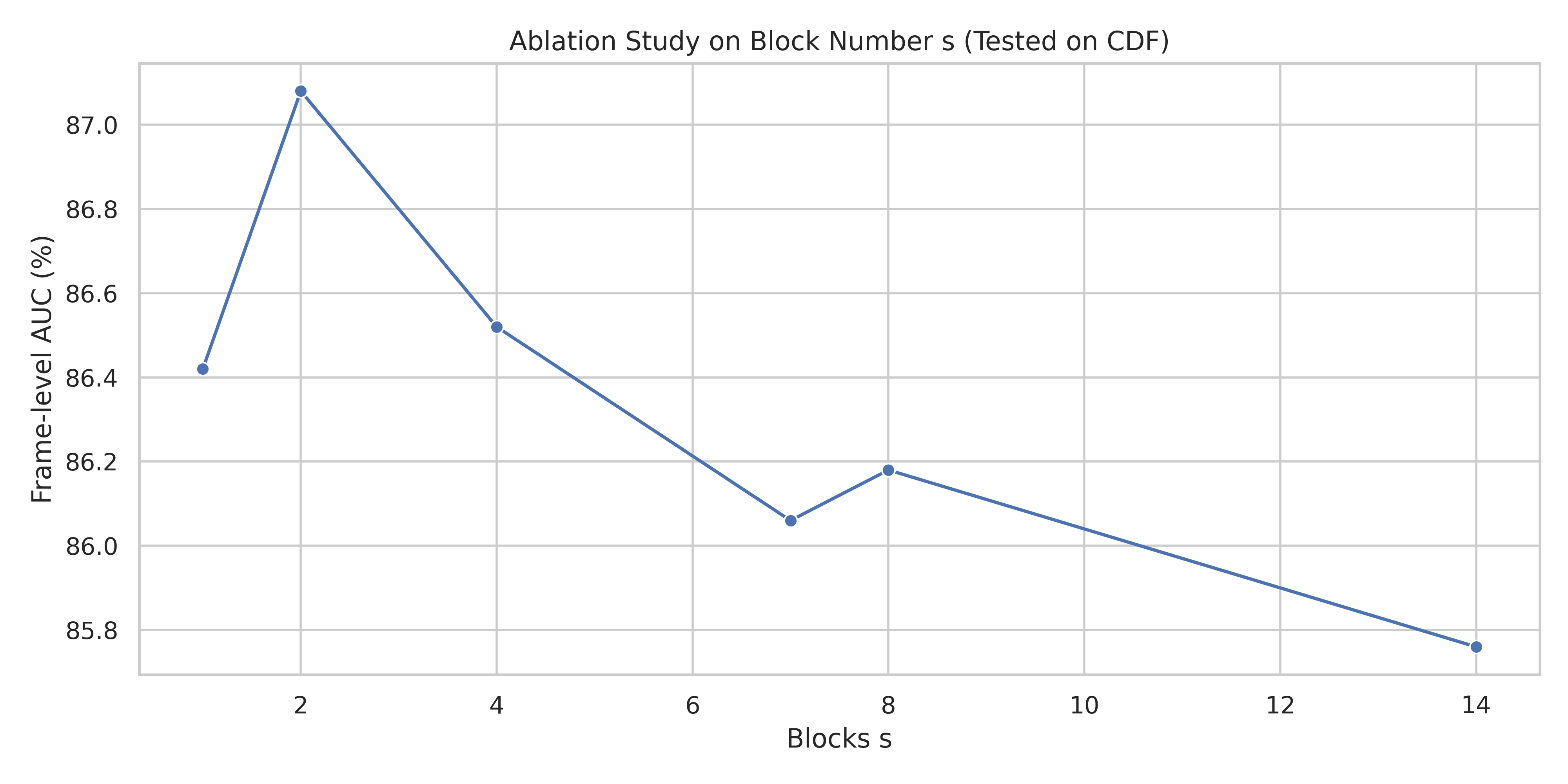}
    \caption{Ablation of the block numbers ($s$). Cross-dataset frame-level AUC (\%) scores are illustrated.}
    \label{fig:blocks}
\end{figure}

We conducted an ablation to show the performance under different numbers of blocks $s \times s$ (see Fig.~\ref{fig:blocks}). 
We see that more blocks (\textit{e.g.,} patchwise shuffling w/ $s=14$) will corrupt the correlation among patches, and the forgery clue (between patches) thus cannot be maintained. 
We find that $2 \times 2$ is the optimal choice in our case. The results suggest that local correlation should be maintained and is relevant to the forgery information and the optimal $s$ is not large.

\subsection{Ablation on Mixing Ratio $\gamma$ in M-Branch}

The selection of mixing ratio $\gamma$ is important as $\gamma$ controls how much content and forgery information are introduced from other samples during mixing. A smaller mixing ratio introduces limited content variance during training and cannot effectively reduce content bias during training. On the other hand, a larger mixing ratio changes the main forgery artifacts within a sample by mixing and breaks the hypothesis for alignment in our framework. As shown in Table~\ref{tab:abl_gamma}, we conduct the experiments with different values of $\gamma$. It turns out that $\gamma = 0.3$ is the best choice among these values (86.74\% AUC on CDF). The performance drops largely (86.74\% $\rightarrow$ 83.93\%) when a large mixing ratio is picked ($\gamma = 0.7$), which meets our expectations. 
Based on the observation, we empirically set $\gamma$ to 0.3 in other experiments.

\subsection{Ablation of the Mixing Module}

Mixing provides \textbf{diverse combinations of content} (\textit{e.g.,} hair, background, ID, and gender) in context and thus reduces the bias, but the ``simple removal" cannot generate more combinations. We compare the cross-dataset performance of removal and the whole mixing module. Results in Tab.~\ref{tab:removal} empirically verify that using removal only leads to a suboptimal overall result in cross-dataset testing. The proposed latent mixing operation effectively enhances the generalization capability of detectors. Therefore, position embeddings that model the position correlation are still important in our case.

\begin{table}
    \centering
    \caption{Comparison between removal (zero out) and mixing.}
    \label{tab:removal}
    
    \begin{tabular}{c|c c c c}
    \toprule
        AUC (\%) &  CDF & DFDCP & DFDC & DFD\\

    \midrule
        Removal & 90.84 & 86.72 & 80.72 & 94.70\\
       Ours  & \textbf{93.13} & \textbf{88.11} & \textbf{81.21} & \textbf{95.51}\\
    \bottomrule
    \end{tabular}
\end{table}

\begin{figure}[!h]
    \centering
    \includegraphics[width=0.75\linewidth]{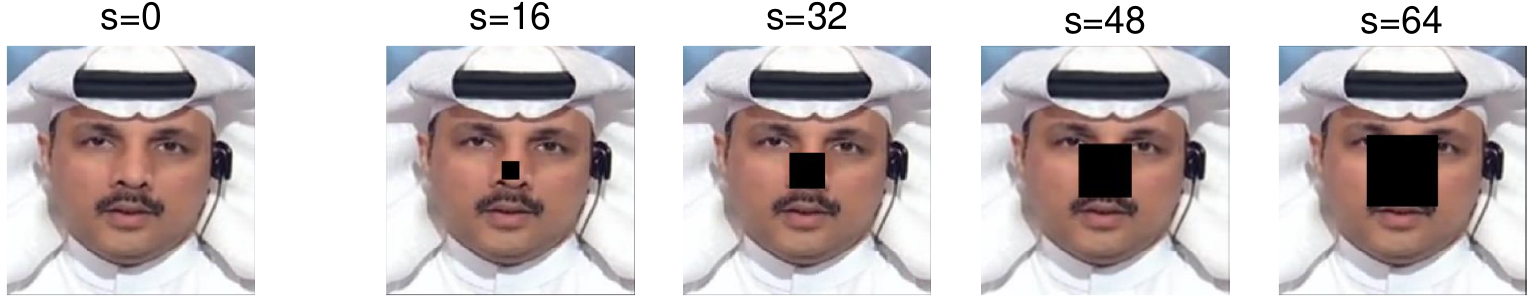}
    \caption{The illustration of images across cutout sizes.}
    \label{fig:cutout}
\end{figure}

\begin{table}[htbp]
    \centering
    \caption{Ablation results on mixing ratio $\gamma$. Frame-level AUC (\%) scores on CDF are reported for comparison.}
    \label{tab:abl_gamma}
    \begin{tabular}{c|c c c c}
        \toprule
       $\gamma$  &  0.1 & 0.3 & 0.5 & 0.7\\
       \midrule
        Ours & 86.14  &\textbf{86.74} & 85.26 & 83.93\\
         \bottomrule
    \end{tabular}
    
\end{table}

\begin{figure*}[tbp]
    \centering
    \includegraphics[width=\linewidth]{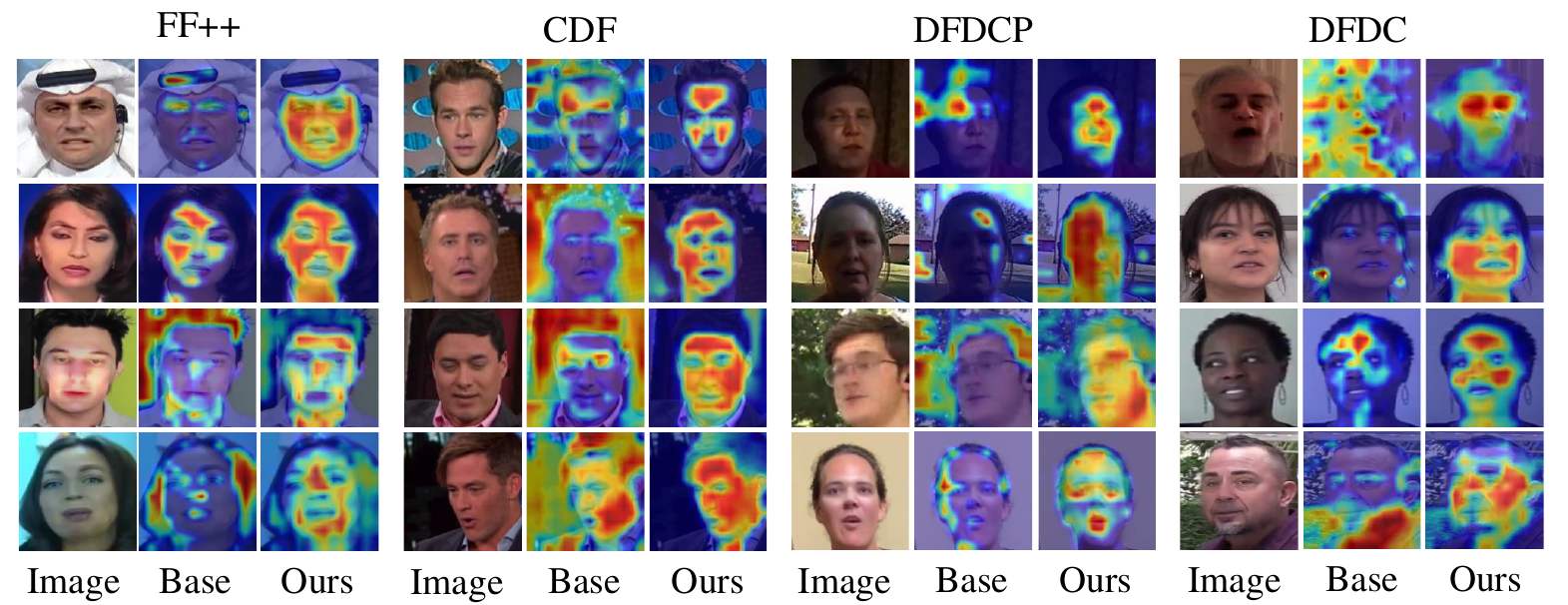}
    \caption{Visualization of heatmaps with GradCAM++. We visualize the heatmaps of the baseline and our model on four datasets FF++, CDF, DFDCP, and DFDC. The results show that the baseline model attends to irrelevant content (hair, clothes, and background) due to the content bias and cannot localize the forgery area due to the position bias. Our model can capture and localize the forgery better from the visualization.}
    \label{fig:gradcam}
\end{figure*}

\section{Quantitative Analysis Bias}

\subsection{Testing with Center-Region Cutout} 
To verify the position bias quantitatively and demonstrate the robustness of our method to the bias, we conduct experiments with the different cutout size $s \times s$ (no cutout when $s=0$) at the center region in Table~\ref{tab:cutout}. 
As shown in Fig.~\ref{fig:cutout}, we illustrate images with different cutout sizes $s \times s$ (no cutout when $s=0$). 
The performance of the baseline model degenerates fast as the cutout size gets larger. Our method achieves more stable performance under different cutout sizes (3.97\% and 1.61\% improvements at cutout sizes of 48 and 64). The results indicate that the baseline model is biased to specific positions (\textit{e.g.}, central region). Our method mitigates this bias and captures richer forgery information, thus performing better under extensive cutout sizes.

\begin{table}[tp]
    \centering
    \caption{Ablation with cutout testing. Frame-level AUC scores on FF++ (c23) are reported for comparison.}
    \label{tab:cutout}
    \begin{adjustbox}{width=\linewidth}
    \begin{tabular}{c|c c c c c}
    \toprule
       Size $s$  & $0$ & $16$ & $32$ & $48$ & $64$ \\
       \midrule
        Baseline & 98.12\% & 97.19\% & 97.63\% & 93.43\% & 95.80\% \\ 
        \multirow{2}{*}{Ours} & 98.66\% & 98.43\% & 98.46\% & 97.40\% & 97.41\% \\ 
        & \textcolor{blue}{($\uparrow 0.54\%$)} & \textcolor{blue}{($\uparrow 1.24\%$)} & \textcolor{blue}{($\uparrow 0.83\%$)} & \textcolor{blue}{($\uparrow 3.97\%$)} & \textcolor{blue}{($\uparrow 1.61\%$)}\\ 
    \bottomrule
    \end{tabular}
    \end{adjustbox}
\end{table}

\begin{table}[htbp]
    \centering
    \caption{More cross-dataset evaluation results. Videl-level AUC (\%) scores are reported.}
    \label{tab:eval}
    \begin{adjustbox}{width=\linewidth}
    \begin{tabular}{c|c c c}
        \toprule
       Methods  & WDF & Fsh & FFIW \\
       \midrule
       Xception~\cite{rossler2019faceforensics++} &66.2 &72.0 & -\\
   FWA~\cite{li2018exposing} & -&65.7 &- \\
       DCL~\cite{sun2022dual} & 77.6 & - & 86.3 \\
       SBI~\cite{shiohara2022detecting} & 70.3 &78.2&84.8\\
       ViT-B (CLIP)~\cite{radford2021learning} & 82.8 & 78.7 & 75.3 \\
       \midrule
        Ours & \textbf{85.2} & \textbf{87.5} & \textbf{89.0}\\
        \bottomrule
    \end{tabular}
    \end{adjustbox}
\end{table}

\begin{figure*}[htbp]
    \centering
    \includegraphics[width=\linewidth]{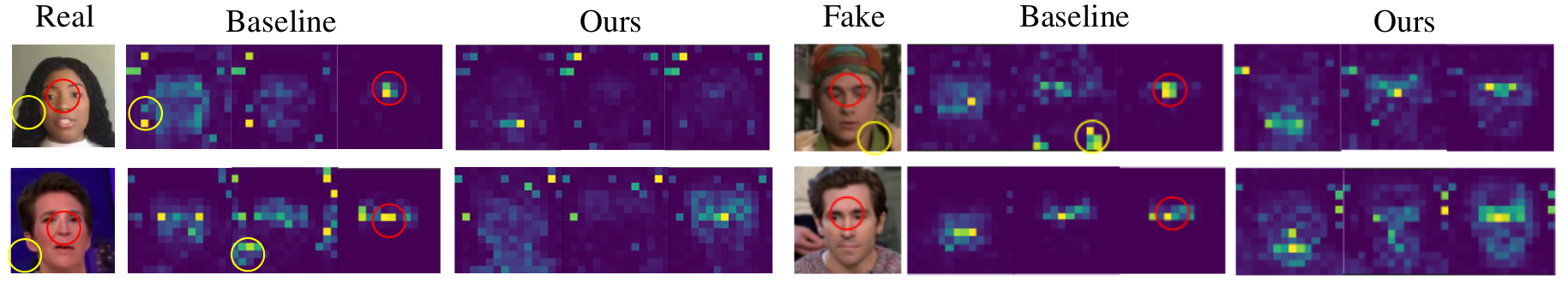}
    \caption{Comparison of attention maps between real and fake faces. \textcolor{red}{Red}: position bias. \textcolor{yellow}{Yellow}: content bias.}
    \label{fig:attn}
\end{figure*}

\begin{figure}
    \centering
    \includegraphics[width=0.8\linewidth]{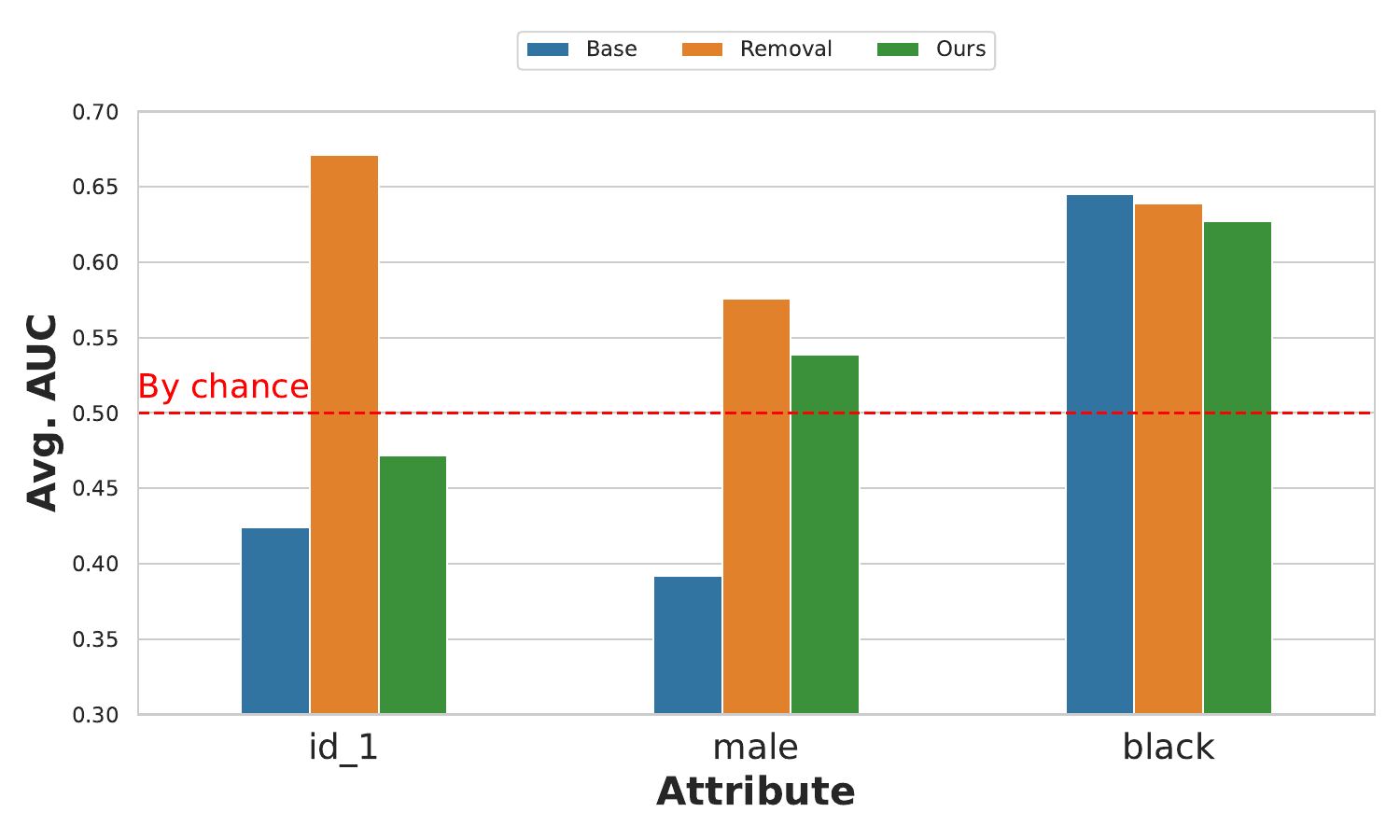}
    \caption{ 
    Classifiers for ``ID, gender, and race" are tested with real/fake labels to assess the impact of the content bias. 
    }
    \label{fig:corr}
\end{figure}

\subsection{Testing with Attribute Classifiers}

To identify the content bias with quantitative analysis, we compare the correlation of classification results between forgery and attributes (\textit{e.g.}, ID, gender, and race) in Fig.~\ref{fig:corr}. ``Removal" follows Sec. B.6 and means simple removal of patches in the mixing module. The attributes represent the content information within an image. We focus on some face-related attributes and first train the attribute linear classifiers with frozen deepfake detector backbones. These classifiers can distinguish the content information within an image. We then test these classifiers with real/fake labels on the FF++/CDF datasets and use averaged AUC scores to analyze the effect of content information on the decision.  The attribute classifiers with our detector backbone achieve AUC scores the most close to 0.5. This means content information contributes little to the forgery decision in our method. The results demonstrate that there exists a correlation between forgery and content for biased detectors (baseline and removal) and our method effectively alleviates the bias towards content information.

\section{Evaluation on More Datasets}

To further demonstrate the generalization capability of the proposed method, we evaluate our method in three additional datasets: WildDeepfake (WDF)~\cite{zi2020wilddeepfake}, FaceShifter (Fsh)~\cite{li2019faceshifter} and FFIW~\cite{zhou2021face}. These datasets are also used for evaluation in some previous works~\cite{sun2022dual, shiohara2022detecting}. It is worth noting that Fsh is an extension of the original FF++~\cite{rossler2019faceforensics++} dataset, in which fake images are produced with a state-of-the-art face swap technique FaceShifter~\cite{li2019faceshifter}. As shown in Table~\ref{tab:eval}, our method outperforms previous methods (including the ViT baseline model) by a large margin and achieves great detection performance on all of the three datasets (85.2\% on WDF, 87.5\% on Fsh and 89.0\% on FFIW).

\begin{figure}[htbp]
    \centering
    \includegraphics[width=0.8\linewidth]{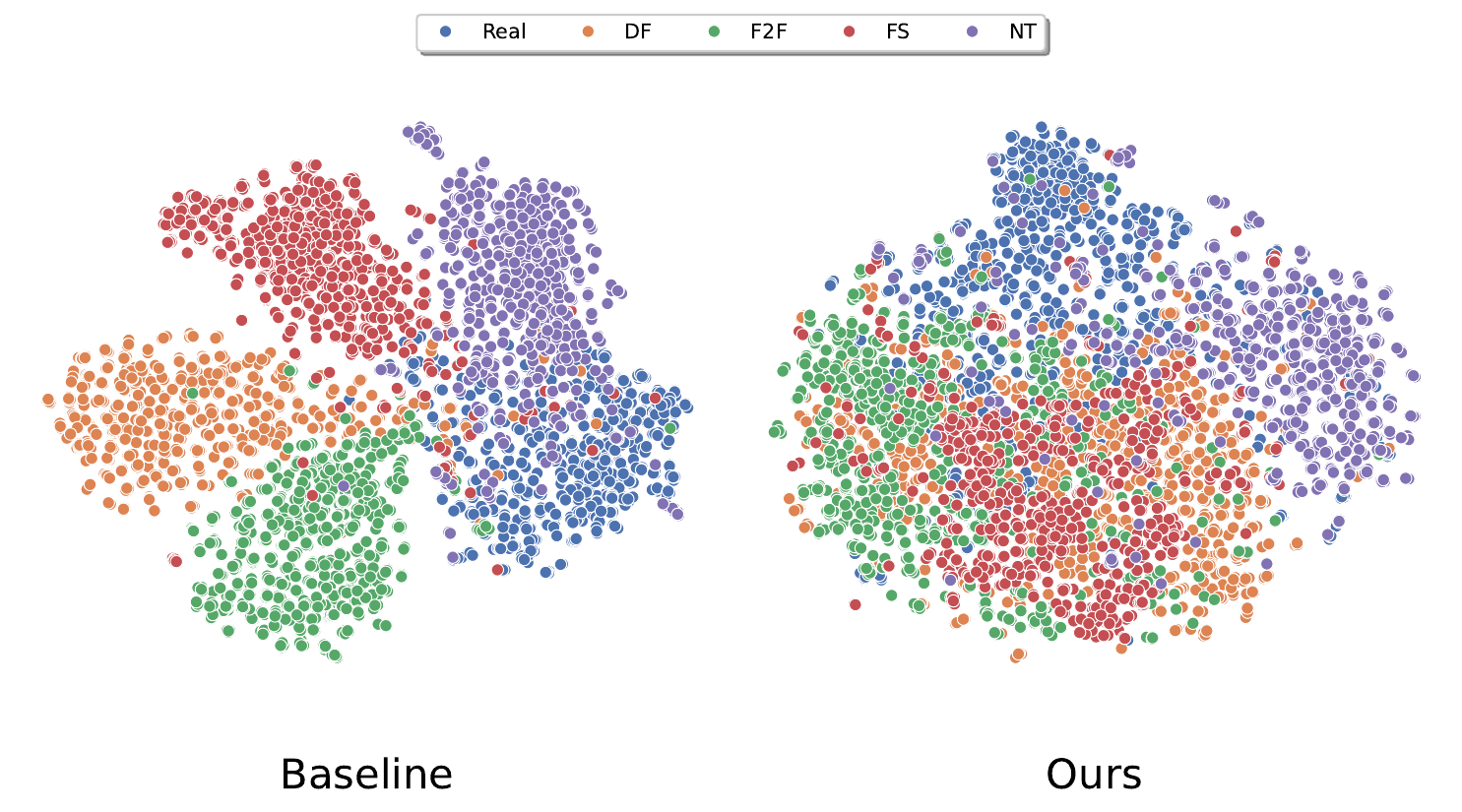}
    \caption{Visualization of t-SNE results. We visualize the feature space for detection on FF++ (c23) and compare our model with the baseline.}
    \label{fig:tsne}
\end{figure}

\section{Visualization}
\subsection{Visualization of T-SNE}

We visualize the feature space of baseline and our models on FF++ (c23) with t-SNE~\cite{van2008visualizing} and analyze the differences. As shown in Fig.~\ref{fig:tsne}, the baseline model separates the features of different forgery methods and suggests the overfitting phenomenon, which is also observed in previous works~\cite{yan2023ucf, zheng2021exploring}. Such a phenomenon indicates that the baseline model may overlook the general forgery features among manipulation methods and thus underperform in generalized deepfake detection. On the contrary, the feature space of our model distinguishes real and fake faces while the connections among manipulation methods are maintained. The results demonstrate that our unbiased learning framework effectively alleviates the overfitting issues during training and implicitly allows the model to capture general forgery artifacts.

\subsection{Visualization of Heatmaps}

To identify whether the model attends to forgery regions within an image, we visualize heatmaps produced by GradCAM++~\cite{selvaraju2017grad,chattopadhay2018grad} on the first block (closely corresponds to the position) of transformer models. As shown in Fig.~\ref{fig:gradcam}, the baseline model attends to irrelevant regions (\textit{e.g.}, hair, clothes, and background) in many cases due to the content bias existing in the training set.  Also, the baseline model attends to specific positions in some cases (\textit{e.g.}, center region in row 3 and row 4 of FF++) but cannot capture the whole forgery area due to the position bias lying in the model. Our method captures the forgery artifacts over the face and neck areas better from heatmaps, which benefit from the explicit reduction of bias in our learning framework.

\subsection{Visualization of Attention Maps}

The attention maps are visualized as in the paper. Here, we compare both real and fake in Fig.~\ref{fig:attn}, where attention maps of fake exhibit \textbf{more activations} than the real (better discrimination). We (w/ our shuffle and mix) also show \textbf{less bias} than the baseline model (w/o ours).

\end{document}